\documentclass{article}

\usepackage[nonatbib,final]{neurips_2023}

\usepackage[utf8]{inputenc} % allow utf-8 input
\usepackage[T1]{fontenc}    % use 8-bit T1 fonts
\usepackage{hyperref}       % hyperlinks
\usepackage{url}            % simple URL typesetting
\usepackage{booktabs}       % professional-quality tables
\usepackage{amsfonts}       % blackboard math symbols
\usepackage{nicefrac}       % compact symbols for 1/2, etc.
\usepackage{microtype}      % microtypography

\usepackage{times}
\usepackage{epsfig}
\usepackage{graphicx}
\usepackage{amsmath}
\usepackage{amssymb}
\usepackage{wrapfig}

\usepackage[ruled,vlined]{algorithm2e}

\usepackage{amsthm}
\usepackage{bbm}

\usepackage{xcolor}
\usepackage{subcaption}
\usepackage{enumitem}
\newcommand{\specialcell}[2][c]{%
  \begin{tabular}[#1]{@{}c@{}}#2\end{tabular}}

\usepackage{wrapfig}

\usepackage{multirow}

\begin{document}

%%%%%%%%% TITLE
\title{Large VLM-based Stylized Sports Captioning}

\author{Sauptik Dhar \and Nicholas Buoncristiani \and Joe Anakata \and Haoyu Zhang \and Michelle Munson\\
Eluvio AI Labs\\
{\tt\small \{sauptik.dhar, nick.buoncristiani, joe.anakata, haoyu.zhang, michelle.munson\}@eluv.io}
}

\maketitle

\textbf{Introduction}  \; The advent of large (visual) language models (LLM / LVLM) have led to a deluge of automated human-like systems in several domains including social media content generation \cite{zeng2024large}, search and recommendation \cite{wu2024surveylargelanguagemodels, search_git}, healthcare prognosis \cite{he2025surveylargelanguagemodels,liu2024surveymedicallargelanguage}, AI assistants\begin{wrapfigure}[12]{r}{0.25\textwidth}
  \begin{center}
   \vspace{-12pt}
    \includegraphics[width=0.25\textwidth]{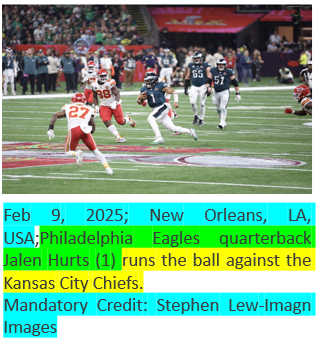}
  \end{center} 
  \vspace{-8pt}
  \caption{Caption Style} \label{caption_style} 
\end{wrapfigure} for cognitive tasks \cite{luo2025largelanguagemodelagent, agent_git}, etc. Although these systems have been successfully integrated in production; very little focus has been placed on sports, particularly accurate identification and natural language description of the game play. Most existing LLM/LVLMs can explain generic sports activities, but lack sufficient domain-centric sports' jargon to create natural (human-like) descriptions. This work highlights the limitations of existing SoTA LLM/LVLMs for generating production-grade sports captions from images in a desired stylized format, and proposes a two-level fine-tuned LVLM pipeline to address that.  The proposed pipeline yields an improvement $>8-10\%$ in the $F_1$, and $>2 - 10\%$ in BERT score compared to alternative approaches. In addition, it has a small runtime memory footprint and fast execution time. During Super Bowl LIX the pipeline proved its practical application for live professional sports journalism; generating highly accurate and stylized captions at the rate of 6 images per 3-5 seconds for over 1000 images during the game play. 

% This work addresses the challenges of generating production-grade captions for sports images in a desired stylized format and summarizes our experiences applying the approach during Super Bowl LIX to caption more than 1000 images during the game. We highlight the limitations of existing SoTA LLM/LVLMs  and propose a two-level fine-tuned LVLM pipeline for generating highly accurate stylized captions. The proposed pipeline yields an improvement $>8-10\%$ in the knowledge density score $F_1$ and a boost in the BERT score $>2 - 10\%$ compared to alternative approaches. The runtime pipeline has a small memory footprint, fast execution, and high-quality output. During Super Bowl 2025 the pipeline proved its practical application for live professional sports journalism, generating captions at a rate of 1 frame/image per 5 seconds, running on commercial off-the-shelf server hosts with large memory GPUs, and it was used to successfully generate over 1,000 usable captions during game play requiring only modest or no human editing.

\textbf{Problem Statement} \; Here the goal is to produce accurate stylized sports captions for professional football in real time. This type of system can significantly reduce production and operating costs for the sports media industry \cite{elv_imagn_news_1, elv_imagn_news_2}. We adopt sports caption style used by Imagn, the largest sports-image wire service in the USA:-  \colorbox{cyan}{Meta Data (Date; Location);}\colorbox{green}{Caption Entity}\colorbox{yellow}{Caption - Action;}\colorbox{cyan}{Meta-Data Credits} (see Fig \ref{caption_style} and \cite{superbowl_imagn}). The images include additional meta-data which is populated by the \begin{wrapfigure}[16]{r}{0.7\textwidth}
  \begin{center}
   \vspace{-10pt}
    \includegraphics[height=1.8in]{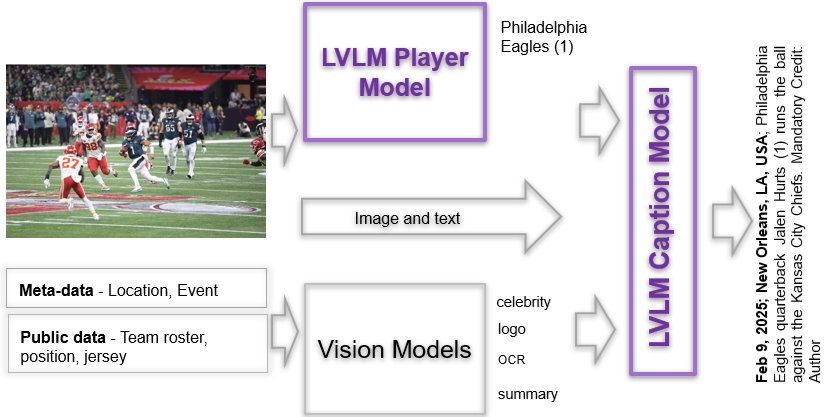}
  \end{center} 
  \vspace{-10pt}
  \caption{Two-level LVLM pipeline for Sports (Football) Caption Generation.} \label{VLM_pipeline} 
\end{wrapfigure}photographer such as the event title, date, location, and credits information. The main task entails identifying the correct entities (players, coach, celebrities, etc.) with their actions and generating the captions in the stylized format. Furthermore, to produce any operational gains, the final solution needs to be resource efficient, i.e. low memory and low latency (at least > 1 image per minute). Our initial qualitative analysis showed that off-the-shelf SoTA LVLMs (zero-shot), or their test-time optimized (few-shot, prompt optimization) versions, lacked the accuracy and style required for production. 

% leveraging off-the-shelf LVLMs (zero-shot) was severely limited in describing images with appropriate sports jargon and in accurately identifying player entities. Modifying such SoTA LVLMs through test time optimizations like few-shot, prompt optimization etc., improved the overall quality of the captions, but was still insufficiently accurate in identifying and describing the plays/players and imitating the style required for production.            

% \begin{figure}[htbp]
%     \centering
%     \includegraphics[height=1.8in]{VLM.png}
%     \caption{Two-level LVLM pipeline for Sports (Football) Caption Generation.} \label{VLM_pipeline}
% \end{figure}

\textbf{Our Approach} \; There are four main tasks in our pipeline: a) accurate entity identification, b) correct association of entity with action, c) correct caption format, and d) resource efficiency. For this, we designed a two-level Large Visual Language Model (LVLM) pipeline as shown in Fig. \ref{VLM_pipeline}. 

\begin{wrapfigure}[14]{r}{0.4\textwidth}
  \begin{center}
   \vspace{-18pt}
    \includegraphics[width=0.4\textwidth]{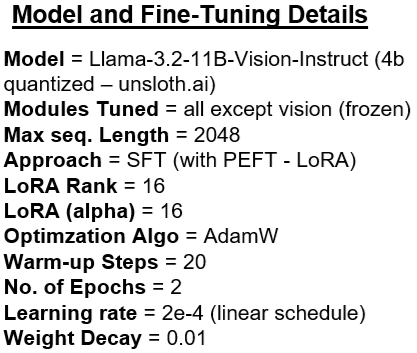}
  \end{center} 
  \vspace{-10pt}
  \caption{SFT Template} \label{sft_template} 
\end{wrapfigure}

\underline{Level 1 - LVLM Player Model} Here we focus on the identification of the player. Note that this is a slightly harder problem than a typical face detection/identification, as in most on-field pictures the players are wearing helmets. Also, a detection / identification based on jersey numbers may be limited due to occlusion. We leveraged Supervised Fine-Tuning (SFT) to build a decoder-based entity (player) recognition model. The SFT template used throughout the paper is provided in Fig. \ref{sft_template}.  The input to the model is the image and its meta-data. We additionally provide the teams' roster as a context. For SFT, we annotate the teams and jersey numbers (say \textit{Philadelphia Eagles (1)}) visible in the images and also add a confidence for the labels as HIGH vs. LOW confidence. Limiting the generation to only the team and jersey numbers reduced the hallucination on player names, which could be deterministically mapped back from the team roster. Further, attributing confidence levels to the labels allowed the LVLM to attend more to the in-focus players, and boosted their True Positive (TP) rate while also reducing the False Positive rate of out-of-focus players. Also, leveraging a decoder as opposed to an encoder-based approach provided additional generalization capability through post-training  optimization mostly for out-of-distribution data. For games with teams' jersey colors changed or team roster changes not covered in training data, post-training prompt optimization provided improved generalization. Finally, training with this modified data labeling provided accuracy boost > 8 - 10\% in $F_{1}$ score.
\begin{wraptable}[10]{r}{0.5\textwidth}
    \centering
    \vspace{-0.5cm}
    \begin{small}
    \caption{Caption quality for Super Bowl LIX \cite{superbowl_imagn}. \\\footnotesize{$^{*}$Worse entity $F_1$ score compared to zero-shot}. \\ \footnotesize{$^{\dagger}$ Finetuned directly on the captions}.} \label{tab:error}
    % \tabcolsep=0.1cm
    
    % \begin{sc}
    \resizebox{0.5\columnwidth}{!}{
    \begin{tabular}{lc}
        \hline 
        \addlinespace
        Method & BERT Score ($\times$ 100) \cite{zhang2020bertscoreevaluatingtextgeneration}\\
        \hline
        \specialcell{\small{Llama3.2 (11B)}-\small{Zero-Shot}} & $81.0 \pm 3.9$ \\
        \specialcell{\small{Llama3.2 (11B)}-\small{Few-Shot}} & $85.2 \pm 2.6 ^{*}$ \\
        \specialcell{\small{Llama3.2 (90B)}-\small{Zero-Shot}} & $84.5 \pm 2.8$ \\
        \specialcell{\small{Llama3.2 (90B)}-\small{Few-Shot}} & $87.1 \pm 2.4 ^{*}$\\
        Finetune - direct{$^{\dagger}$}& $89.8 \pm 2.1 $ \\
        \specialcell{\small{\textbf{Finetune (ours)}}-\small{Two-Level}}& $91.2 \pm 2.8$ \\
       %   Method & BERT Score \cite{}\\
       %   \hline
       %   \specialcell{\small{Llama3.2(11B)}\\\small{Zero-Shot}} & $22.86$ \\
       %   \specialcell{\small{Llama3.2(11B)}\\\small{Zero-Shot}} & $16.61 \pm 0.24$ \\
       %   \specialcell{\small{Llama3.2(11B)}\\\small{Zero-Shot}}& $19.65 \pm 1.74$  \\
       %    \specialcell{\small{Llama3.2(11B)}\\\small{Zero-Shot}} & $19.65 \pm 1.74$  \\
       %     Textgrad& $19.65 \pm 1.74$  \\
       %  Finetune - direct $\dagger$& $19.65 \pm 1.74$  \\
       % \specialcell{\small{Finetune (ours)}\\\small{Two-Level}}& $19.65 \pm 1.74$  \\
       % & $19.65 \pm 1.74$  \\
         \hline
    \end{tabular}
    }
    \end{small}
    \end{wraptable}
    
\underline{Level 2 - LVLM Caption Model} \; Next we built a fine-tuned LVLM for correct identification of actions and their association with the right entities (players). This LVLM is also tasked with learning the desired caption style. The input to the model includes the image and its meta-data, the teams' roster, output from LVLM player model, and output from several (traditional) vision-based models which identifies: celebrity faces, logos, OCR, generic image description (using \cite{wang2022gitgenerativeimagetotexttransformer}). For SFT, we re-annotated the ground-truth captions by retaining only the HIGH-confidence players identified in the Level 1 annotation. This helped the LVLM to attend more to the players in focus and reduce hallucinations. The use of preferential alignment with ORPO \cite{hong2024orpomonolithicpreferenceoptimization} did not yield significant improvements and is therefore not reported. The final pipeline provided a significant boost in the overall quality of the captions (see Table \ref{tab:error}). Compared to a direct fine-tuning model, the proposed approach also provided better accuracy $F_1 \text{score} >8\%$. Also (see Fig. \ref{sft_template}) the final models are quantized in 4 bits, which provides significant VRAM efficiency ($>10\times$ compared to 11B and $>90 \times$ compared to 90B) during inference. Although there was no significant improvement in computation speed compared to test-time optimized models (such as zero-shot, few-shot, etc.), the final latency of the system satisfied the operational cost constraints by $>72 \times$ the required margin.

\textbf{Summary} \; In this work we built a production grade sports caption generator and used it to successfully generate over 1000 usable captions during the Super Bowl 2025 game play. Our proposed approach provided significant accuracy gains and captured the appropriate sports jargon in the desired caption format. Finally, it achieved the required latency constraints by >72$\times$ while providing a significant reduction in run-time memory.  

% We use football (Super Bowl LIX) as an example and propose a two-level fine-tuned LVLM model pyramid for accurate and stylized captions. The proposed approach provided significant \textit{accuracy gains} and captured the appropriate \textit{ sports jargon} in the \textit{desired caption format}. Finally, the fine-tuned models achieved the required latency constraints by >100$\times$ while providing a significant reduction in inference memory footprint.   

\textbf{Future Research} While the decoder-based approach provided significant advantage for the Level 1 player model for out-of-distribution images, a thorough research on their advantages compared to encoder-based ones for visual/video LLMs is an ongoing research. Finally, scalable and accurate hyperparameter tuning of LVLMs (under SFT or ORPO settings) is nontrivial and showed high variability in fine - tuned performance. This is still on-going research.    

\textbf{Acknowledgment} We thank the entire Imagn team led by Bruce Odle for the collaboration. In addition, we thank Arun Antonio and Victor Amigo for the help with data annotation.

{\small
\bibliographystyle{ieee_fullname}
\bibliography{VLMsports}
}

\end{document}